\documentclass[11pt]{article}

\usepackage{ACL2023}

\usepackage{booktabs}
\usepackage{amsmath}
\usepackage{bm}
\usepackage{multirow}
\usepackage{lineno}
\usepackage{array}
\usepackage{mathtools}

\usepackage{times}
\usepackage{latexsym}

\usepackage[T1]{fontenc}

\usepackage[utf8]{inputenc}

\usepackage{microtype}

\usepackage{inconsolata}

\usepackage{hyperref}
\usepackage{url}

\usepackage{wrapfig}        
\usepackage[tight,footnotesize]{subfigure}
\usepackage{caption}
\usepackage{multirow,multicol}
\usepackage{algpseudocode}
\usepackage{amsmath,amssymb,amsfonts,graphicx,amsthm}
\usepackage{color,soul,xspace}
\usepackage{booktabs}       
\usepackage{physics}        
\usepackage{colortbl}  

\usepackage{enumitem}
\usepackage{bookmark}
\usepackage{bm}
\usepackage{amsfonts}
\usepackage{subfigure}
\usepackage{graphicx}
\usepackage{makecell}
\usepackage{xcolor}
\usepackage{arydshln}
\usepackage{stmaryrd}
\usepackage{makecell}
\usepackage{paralist}

\newcommand{\eat}[1]{}

\definecolor{Gray}{gray}{0.95}
\definecolor{Cyan}{rgb}{0.88,1,1}
\definecolor{Blu}{RGB}{68,115,196}

\newcommand{\todo}[1]{\textcolor{red}{[TODO: #1]}} 
\newcommand{\zfq}[1]{\textcolor{red}{zfq: #1}}

\newcolumntype{a}{>{\columncolor{Gray}}c}

\newcommand{\vsa}{\vspace*{-0.28cm}}

\usepackage{rotating}
\usepackage{adjustbox}

\title{A Diffusion Model for Event Skeleton Generation}

\author{
    Fangqi Zhu\textsuperscript{\rm 1,\rm 3}\footnotemark[1],
    Lin Zhang \textsuperscript{\rm 3},
    Jun Gao\textsuperscript{\rm 1},
    Bing Qin\textsuperscript{\rm 1},   
    Ruifeng Xu\textsuperscript{\rm 1,\rm 2}\footnotemark[2], 
    Haiqin Yang\textsuperscript{\rm 3}\footnotemark[2] \\
    \textsuperscript{\rm 1} \normalsize Harbin Institute of Technology, Shenzhen, China \\
    \textsuperscript{\rm 2} \normalsize Guangdong Provincial Key Laboratory of Novel Security Intelligence Technologies \\
    \textsuperscript{\rm 3} \normalsize International Digital Economy Academy (IDEA) \\
    \texttt{zhufangqi.hitsz@gmail.com, xuruifeng@hit.edu.cn, hqyang@ieee.org}
}

\begin{document}

\maketitle
\renewcommand{\thefootnote}{\fnsymbol{footnote}} 
\footnotetext[1]{Work done when Fangqi was interned at IDEA.  } 
\footnotetext[2]{Corresponding authors.} 
\renewcommand{\thefootnote}{\arabic{footnote}}

\begin{abstract}
    Event skeleton generation, aiming to induce an event schema skeleton graph with abstracted event nodes and their temporal relations from a set of event instance graphs, is a critical step in the temporal complex event schema induction task.
    Existing methods effectively address this task from a graph generation perspective but suffer from noise-sensitive and error accumulation, e.g., the inability to correct errors while generating schema.  
    We, therefore, propose a novel Diffusion Event Graph Model~(DEGM) to address these issues. 
    Our DEGM is the first workable diffusion model for event skeleton generation, where the embedding and rounding techniques with a custom edge-based loss are introduced to transform a discrete event graph into learnable latent representation. 
    Furthermore, we propose a denoising training process to maintain the model's robustness.
    Consequently, DEGM derives the final schema, where error correction is guaranteed by iteratively refining the latent representation during the schema generation process. Experimental results on three IED bombing datasets demonstrate that our DEGM achieves better results than other state-of-the-art baselines. Our code and data are available at \url{https://github.com/zhufq00/EventSkeletonGeneration}.
\end{abstract}

\section{Introduction}

\begin{figure}[!t] 
    \centering
    \includegraphics[width=0.8\linewidth]{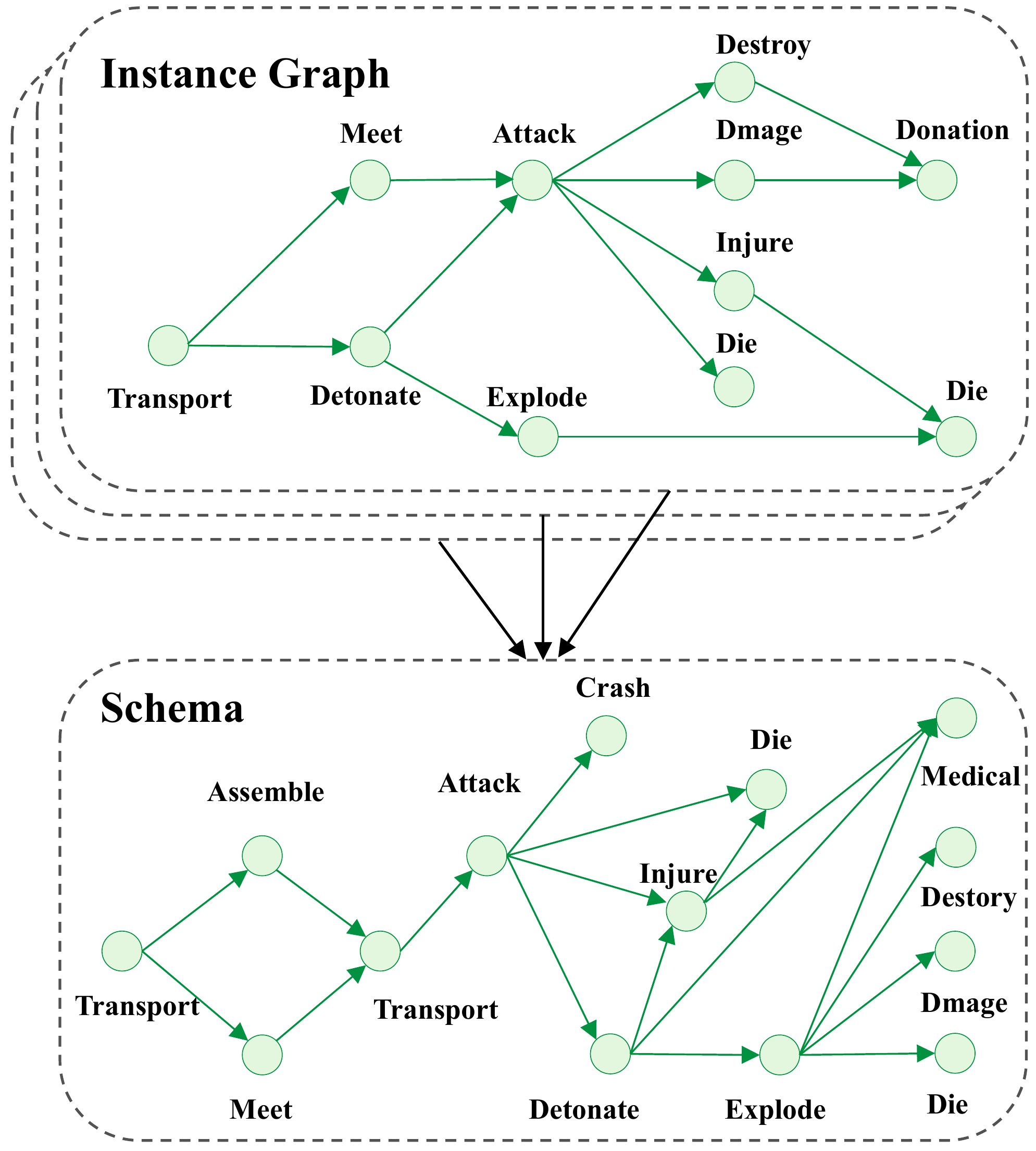}
    \caption{
    An illustrated example demonstrates the utilization of multiple instance graphs extracted from news articles depicting complex events to generate an event schema skeleton graph for the complex event type \textit{Car bombing}. The presented instance graph specifically represents the complex event known as the \textit{Kabul ambulance bombing}. A circle symbolizes an atomic event.} 
    

    \label{fig:intro}
\end{figure}

Event schema induction is to identify common patterns and structures in event data, which can extract high-level representation of the events. 
Current event schema induction tasks mainly focus on simple event schemas, e.g.,  templates~\citep{chambers-2013-event} and scripts~\citep{chambers-jurafsky-2009-unsupervised}.  However, real-world events are usually more complex, which include multiple atomic events, entities, and their relations, which require more advanced techniques to adequately capture and represent the different aspects and relations involved. 

Recently, \citet{li-etal-2021-future} propose the temporal complex event schema induction task in order to understand these complex events. The task seeks to abstract a general evolution pattern for complex events from multiple event instance graphs.  It is divided into two subtasks: event skeleton generation and entity-entity relation completion. The first task focuses on creating the event skeleton, i.e., representing each atomic event with its associated event type as an event node and exploring their temporal relations.  The second one is to complete entities and entity links for the event skeleton.  In this paper, we focus on event skeleton generation as it is a prerequisite yet formidable task in temporal complex event schema induction. Figure~\ref{fig:intro} illustrates an example of instance graphs\footnote{For simplicity, we mention ``schema'' as ``event schema skeleton graph'', ``instance graph'' as ``event instance skeleton graph'', and ``event graph'' represents both.} and the corresponding abstracted schema.  Both include abstract event types, such as \textit{Attack}, and their temporal relations, like \textit{Injure} happening after \textit{Attack}.  

Event skeleton generation requires a deep understanding of events and their multi-dimensional relations. 
Previous methods employ autoregressive graph generation models to generate a schema, sequentially generating event nodes from the previous ones.
For example, \citet{li-etal-2021-future} generate the event node with its potential arguments and propagates edge-aware information within the temporal orders.
 \citet{jinetal2022event} improves this approach by applying a Graph Convolutional Network (GCN) to better capture structural information in instance graphs and adopting a similar autoregressive generation approach to generate event graphs.  However, autoregressive generation methods for event skeleton generation result in errors accumulating over time, which may degrade the performance of the generation model. For instance, as shown in Figure \ref{fig:intro}, the model may mistakenly generate ``Explode'' as ``Die'', causing it to fail to generate subsequent events correctly.
 Intuitively, as the number of event nodes increases, the error accumulation becomes more severe. This comes from two factors.  The first one is error propagation in the autoregressive graph generation models because they are noise-sensitive and strongly rely on the correctness of the generated node.  If the model generates an incorrect node, it will lead to a cascading effect of errors in generating the schema.  Robustness is a serious issue in autoregressive methods. The second factor is the model's inability to correct errors in the generation procedure.  Hence, we need a model, which can correct the generated event-type nodes during generating.

To this end, we propose a novel event graph generation model, dubbed Diffusion Event Graph Model~(DEGM), to address these issues. 
To battle the model's robustness, we propose a diffusion-based method, inspired by the outstanding performance in recent research~\cite{sun2022pointdp,xiao2022densepure}. By carefully selecting the amount of Gaussian noise in the diffusion process, the model can remove adversarial perturbations, thereby increasing the model's robustness. However, there are still two challenges in applying this method directly to the event graph: (1) mapping the discrete graph structures and event types to a continuous space, and (2) finding a way to recover the event graph from the continuous space.
We then develop the denoising stage, including converting the event graph into a sequence and applying an embedding technique to project it to the continuous space. Additionally, we introduce a custom edge-based loss function to capture the missing structural information during the transformation.  To tackle the second challenge, we develop a rounding technique to predict the event types based on their representation and a pre-trained classifier to predict the event edges. 
To address the second issue, we derive the final schema, which guarantees error correction, by iteratively refining the latent representation.
We summarize our contributions as follows: 
\begin{compactitem}
    \item We propose a novel Diffusion Event Graph model~(DEGM) for event skeleton generation, in which a denoising training stage guarantees the model's robustness and the schema generation process fulfills error correction via iterative refinement on the latent representation.
    \item We are the first to tackle event skeleton generation via diffusion models, where we convert an event graph from discrete nodes to latent variables in a continuous space and train the model parameters by optimizing the event sequence reconstruction and graph structure reconstruction simultaneously. 
    \item Experimental results on the event skeleton generation task demonstrate that our approach achieves better results than state-of-the-art baselines.
\end{compactitem}
\section{Preliminaries and Problem Statement}
\subsection{ Diffusion Models in a Continuous Space}
A diffusion model typically consists of forward and reverse processes. Given data $\mathbf{x}_0\in\mathbb{R}^d$, the forward process gradually adds noise to $\mathbf{x}_0$ to obtain a sequence of latent variables in $\mathbb{R}^d$, $\mathbf{x}_1, \dots, \mathbf{x}_T$, where $\mathbf{x}_T$ is a Gaussian noise.  Formally, the forward process can be attained by  $q\left(\mathbf{x}_t \mid \mathbf{x}_{t-1}\right)=\mathcal{N}\left(\mathbf{x}_t ; \sqrt{1-\beta_t} \mathbf{x}_{t-1}, \beta_t \mathbf{I}\right)$, where $\beta_t$ controls the noise level at the  $t$-th step. Denote $\alpha_t = 1-\beta_t$ and $\overline{\alpha}_t=\sum_{s=1}^{t}\alpha_s$, we can directly obtain $\mathbf{x}_t$ as $q\left(\mathbf{x}_t \mid \mathbf{x}_{0}\right)=\mathcal{N}\left(\sqrt{\overline{\alpha}_t} \mathbf{x}_{0}, 1-\overline{\alpha}_t \mathbf{I}\right)$. After the forward process is completed, the reverse denoising process can be formulated as $p_\theta\left(\mathbf{x}_{t-1} \mid \mathbf{x}_t\right)=\mathcal{N}\left(\mathbf{x}_{t-1}; \mu_\theta\left(\mathbf{x}_t, t\right), \Sigma_\theta\left(\mathbf{x}_t, t\right)\right)$ where $\mu_\theta(\cdot)$ and $\Sigma_\theta(\cdot)$ can be implemented using a neural network.
\subsection{ Diffusion Models in a Discrete Space}
For discrete data, e.g., text, \citet{li2022diffusion} employ embedding and rounding techniques to map the text to a continuous space, which can also be recovered. 

Given the embedding of the text $\mathbf{w}$, $\textsc{EMB}(\mathbf{w})$, and suppose $\mathbf{x}_0$ is computed as $q(\mathbf{x}_0|\mathbf{w})=\mathcal{N}(\mathbf{x}_0;\mathbf{w},\beta_0\mathbf{I})$, the corresponding training objective is
\newcommand{\exper}[1]{\textsc{#1}}
\newcommand{\expmb}[1]{\mathbf{#1}}
\newcommand{\expobj}[1]{\mathcal{L}_\text{#1}}
\newcommand{\xx}{\expmb{x}}
\newcommand{\alphabar}{\bar{\alpha}}
\newcommand{\ww}{\expmb{w}}
\newcommand{\cc}{\expmb{c}}
\newcommand{\cx}[1]{\mathbf{x}_{#1}}
\newcommand{\ptheta}{p_\theta}
\newcommand{\mean}{\text{mean}}
\newcommand{\ptrain}{p_\text{train}}
\newcommand{\plm}{p_\text{lm}}
\newcommand{\qphi}{q_\phi}
\newcommand{\Lsimple}{\expobj{simple}}
\newcommand{\Lvlb}{\expobj{vlb}}
\newcommand{\Lsimpleee}{\mathcal{L}^\text{e2e}_\text{simple}}
\newcommand{\xstart}{\xx^{(0)}}
\newcommand{\LookUp}{\exper{LookUp}}
\newcommand{\LM}{\exper{LM}}
\newcommand{\DiffLM}{\exper{Diffusion-LM}}
\newcommand{\LMphi}{\exper{LM}_\phi}
\newcommand{\sep}{\exper{SEP}}
\newcommand{\prefix}{\exper{Prefix}}
\newcommand{\adapter}{\exper{Adapter}}
\newcommand{\FTtop}{\exper{FT-top2}}
\newcommand{\Emb}{\exper{Emb}}
\newcommand{\FT}{\exper{FT-full}}
\newcommand{\infix}{\exper{Infix}}
\newcommand{\MLP}{\exper{MLP}}
\newcommand{\yy}{\textsf{Y$_{\text{idx}}$}}
\newcommand{\pp}{\textsf{P$_{\text{idx}}$}}
\newcommand{\ppp}{\textsf{P'}}
\newcommand{\PrefixMatrix}{P}
\newcommand{\Ebare}[1][]{\mathop{\mathbb{E}}_{{#1}}}
\begin{equation}
\small
\begin{aligned}
& \mathcal{L}^\text{e2e}_{\cx{0}\text{-simple}} (\ww) = \mathop{\mathbb{E}}_{\qphi(\cx{0:T} | \ww)}  \left[\sum_{t=2}^T [||\cx{0}-f_{\theta}(\cx{t}, t) ||^2] \right] +\\
& \mathop{\mathbb{E}}_{\qphi(\cx{0:1} | \ww)} \left[|| \Emb(\ww) - f_{\theta}(\cx{1}, 1) ||^2  - \log \ptheta(\ww | \cx{0})\right].
\end{aligned}
\label{eq:l_e2e}
\end{equation}

The first expectation is to train the predicted model $f_{\theta}(\cx{t}, t)$ to fit $\mathbf{x}_0$ from $2$ to $T$.  Empirically, it can  effectively reduce rounding errors~\citep{li2022diffusion}.  The second expectation consists of two terms: the first item makes the predicted $\mathbf{x}_0$, i.e., $f_{\theta}(\cx{1}, 1)$, closer to the embedding $\textsc{EMB}(\mathbf{w})$ while the second item aims to correctly round $\mathbf{x}_0$ to the text $\mathbf{w}$.

\subsection{Problem Statement}
Event skeleton generation is a subtask of temporal complex event schema induction~\cite{li-etal-2021-future}.  It aims to automatically induce a schema from instance graphs for a given complex event type, where a complex event type encompasses multiple complex events; see an example of \textit{car-bombing} shown in Fig.~\ref{fig:intro}. An event schema skeleton consists of nodes for atomic event types and edges for their temporal relations.  Since event skeleton generation is a prerequisite yet challenging task in the temporal complex event schema induction task, we focus on this task in our work.  


Formally, let $G=(\mathcal{N},\mathcal{E})$ be an instance graph with $N=|\mathcal{N}|$ nodes in $\mathcal{N}$ and $\mathcal{E}$ be the set of directed edges, one can obtain the corresponding adjacency matrix, $\mathbf{A}=\left\{a_{ij}\right\}\in\left\{0,1\right\}^{N\times N}$, where $a_{ij}=1$ if $edge(i,j)\in \mathcal{E}$ and $a_{ij}=0$ otherwise.  Due to temporal relations, $G$ is a directed acyclic graph~(DAG), and $\mathbf{A}$ is an upper triangular matrix.
Each node $n\in \mathcal{N}$ represents an atomic event and is assigned with an event type $n_e\in{\Phi}$, where $\Phi$ denotes the set of event types.  The type of each atomic event is abstracted by the DARPA KAIROS ontology~\footnote{\url{https://nlp.jhu.edu/schemas/}} based on its event mention.   In practice, we extract a set of instance graphs $\mathcal{G}$ as outlined in Sec.~\ref{datasets} from news articles, where each instance graph $G\in\mathcal{G}$ describes a complex event, e.g., \textit{Kabul ambulance bombing} as shown Fig.~\ref{fig:intro}.
Given an instance graph set $\mathcal{G}= \left\{G_1, G_2, \cdots\right\}$, our goal is to generate a schema $S$ that outlines the underlying evolution pattern of complex events under the given complex event type. 

\begin{figure*}[!ht] 
    \centering
    \includegraphics[width=0.95\linewidth]{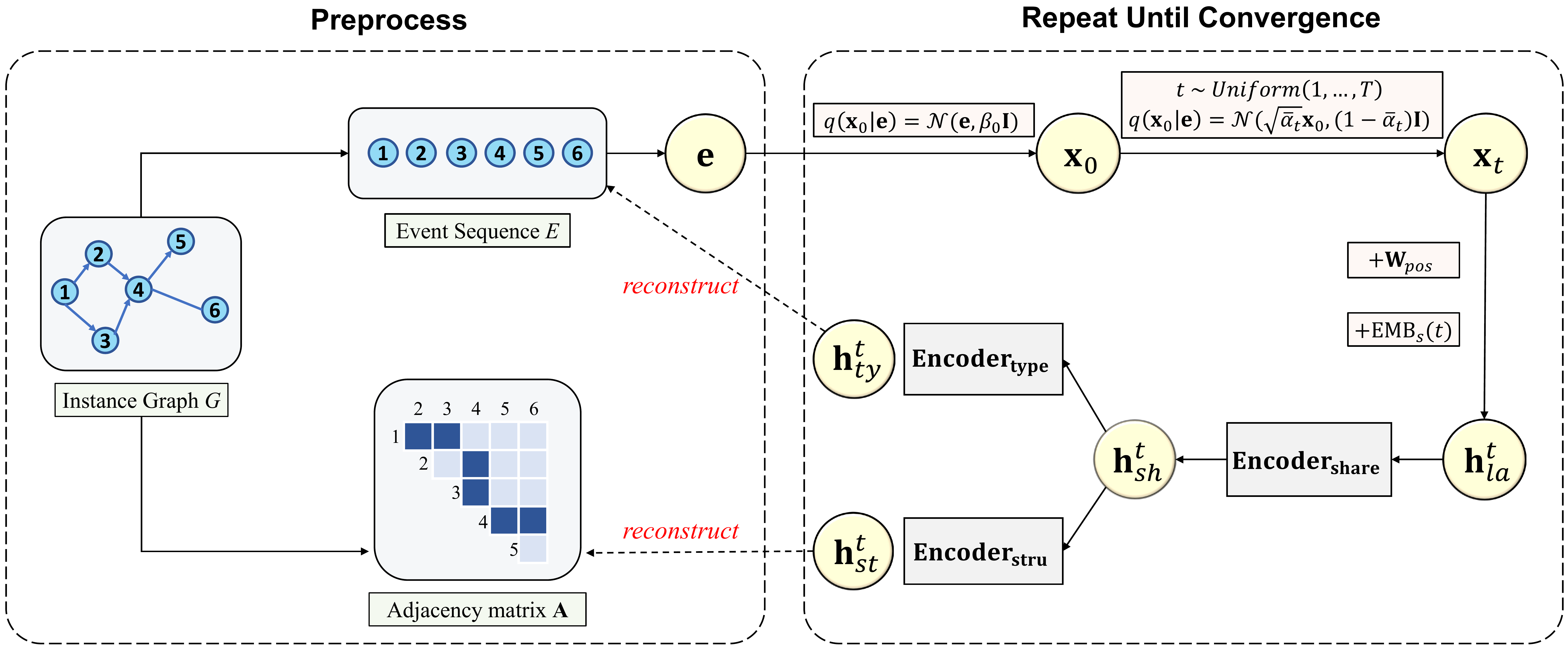}
    \caption{
    The procedure of training our DEGM.  At the preprocessing step, an instance graph $G$ is converted into a temporal sequence of events $\mathbf{e}$ via topological sorting and the associated adjacency matrix $\mathbf{A}$, which represents the graph structure.  Following that, we perform DEGM accordingly. We first convert the discrete events into their representation in a continuous space.  The forward step and the denoising step are conducted iteratively to reconstruct the event sequence and the graph structure.  Note that we convert the latent variable $\mathbf{h}^t_{la}$ into three representation in two levels, i.e., the shared representation $\mathbf{h}^t_{sh}$ and two task-specific representation for the node's type $\mathbf{h}^t_{ty}$ and the node's structure $\mathbf{h}^t_{st}$, respectively; see more details in the text.
    }
    \label{fig:model}
\end{figure*}


\section{Method}
We propose Diffusion Event Graph Model~(DEGM) to tackle the event skeleton generation task. Our DEGM is capable of generating temporal event graphs from random noise.  Fig.~\ref{fig:model} illustrates an overview of our DEGM.

\subsection{Denoising Training}
\label{sec:training}
The denoising training stage consists of three steps to reconstruct the event sequence and graph structure: 1) mapping the event graph into its {\bf embedding representation} in a continuous space; 2) performing a {\bf forward step} to obtain the latent variables, or representation with various levels of noise; 3) conducting the {\bf denoising step} to remove the introduced noise from latent representation.


\paragraph{Embedding representation}
Given an instance graph $G$, we first convert it into a sequence of $m$ events, $E=[e_1, e_2, \ldots, e_m]$, where $e_i$ denotes the event type of node $i$, via topological sorting. We then project $E$ into its embedding representation in a continuous embedding space, 
\begin{equation}
    \mathbf{e}=[\textsc{EMB}_{e}(e_1), \dots,\textsc{EMB}_{e}(e_m)]\in \mathbb{R}^{d\times m},
\end{equation}
where $d$ is the representation size.  Note that $m$ is a preset number of nodes to ensure all graphs are well-aligned.  For graphs with less than $m$ nodes, we pad them by a pre-defined event type: \textsc{PAD}, which makes the total number of event types, $M=\left|\Phi\right|+1$.



\paragraph{Forward Step} 
After obtaining the embedded event sequence $\mathbf{e}$, we deliver the forward process in the diffusion framework to acquire a sequence of latent variables by monotonically increasing the level of introduced noise. 
We sample variables of $\mathbf{x}_0$ and $\mathbf{x}_t$ via 
\begin{align}
\footnotesize
q(\mathbf{x}_0|\mathbf{e})& =\mathcal{N}(\mathbf{x}_0; \mathbf{e},\beta_0\mathbf{I}), \\    q(\mathbf{x}_t|\mathbf{x}_{0})&=\mathcal{N}(\mathbf{x}_t; \sqrt{\overline{\alpha}_t}\mathbf{x}_{0},(1-\overline{\alpha}_t)\mathbf{I}),
\end{align}
where $t=1, \ldots, T$.  Moreover, we introduce two additional embeddings to enhance the expressiveness of latent variables, i.e., the absolute position embedding $\mathbf{W}_{pos}\in \mathbb{R}^{m\times d}$ and the step embedding $\textsc{EMB}_{s}(t)$.  They allow us to capture the event's temporal order in the obtained event sequence and specify that it is at the $t$-th diffusion step.
Adding them together, we obtain the latent variables at $t$-th diffusion step as 
\begin{equation}   \mathbf{h}^t_{la}=\mathbf{x}_t+\mathbf{W}_{pos}+\textsc{EMB}_{s}(t).
\end{equation}

\paragraph{Denoising Step}
Before optimizing the two objectives, event sequence reconstruction and graph structure reconstruction, we first convert the latent variable $\mathbf{h}^t_{la}$ into three variables in two levels, i.e., via a shared encoder $\mathrm{E}_{sh}$ to $\mathbf{h}^t_{sh}$ and two task-specific encoders, the node's type encoder $\mathrm{E}_{ty}$ to $\mathbf{h}^t_{ty}$ and the node's structure encoder $\mathrm{E}_{st}$ to $\mathbf{h}^t_{st}$.  That is, 
\begin{align}
\mathbf{h}^t_{sh} = \mathrm{E}_{sh}(\mathbf{h}^t_{la}), \\
\mathbf{h}^t_{ty} = \mathrm{E}_{ty}(\mathbf{h}^t_{sh}), \\ \mathbf{h}^t_{st} = \mathrm{E}_{st}(\mathbf{h}^t_{sh}).
\end{align}


In the following, we outline the procedure of constructing encoders $\mathrm{E}_{sh}$, $\mathrm{E}_{ty}$, and $\mathrm{E}_{st}$, each contains $l$ layer.  With a little abuse of notations, we define $\mathbf{h}=[\mathbf{h}_1, \dots, \mathbf{h}_m]$ as the input representation for a layer and the corresponding output as $\mathbf{h}^{'}=[\mathbf{h}^{'}_1, \dots, \mathbf{h}^{'}_m]$.

Here, we utilize the graph-attention~\cite{veli2018graph} to transform the input representation into a high-level representation as follows: \vsa 
\begin{equation}
\mathbf{h}^{'}_{i}=\sigma \big(\sum \limits_{j=1}^{m}\alpha_{ij}\mathbf{W}\mathbf{h}_j \big),
\end{equation}
where $\mathbf{W}\in \mathbb{R}^{d\times d}$ is a weight matrix, $\sigma(\cdot)$ is a nonlinear activation function.  Here, $\alpha_{ij}$ is the attention weight defined by 
\begin{equation}
\footnotesize
    \alpha_{ij}=\frac{\mathop{\exp}\big(\mathrm{LR}(\mathbf{a}^T[\mathbf{W}\mathbf{h}_i\|\mathbf{W}\mathbf{h}_j])\big)}{\sum \limits_{k=1}^{m}\mathop{\exp}\big(\mathrm{LR}(\mathbf{a}^T[\mathbf{W}\mathbf{h}_i\|\mathbf{W}\mathbf{h}_k])\big)},
\end{equation}
where $\mathbf{a}\in \mathbb{R}^{2d}$ is a weight vector, $\mathrm{LR}$ is the LeakyReLU activation function, and $||$ denotes the concatenation operation.  We compute attention weights in this way instead of relying on the inner product to prevent higher attention weights between atomic events of the same event type~\footnote{\citet{wu2022nodeformer} observe that using the inner product to calculate attention weights results in higher weights between nodes of the same type.}, which is not appropriate for constructing the event graph.  For instance, the attention weight between two independent \textit{Attack} events should be less than the weight of one \textit{Attack} and its successor events.
After attaining $\mathbf{h}^t_{ty}$,$\mathbf{h}^t_{st}$, via $\mathrm{E}_{ty}$ and $\mathrm{E}_{st}$, respectively, we compute two losses, the event sequence reconstruction loss $\mathcal{L}^{t}_{ty}(G)$ and the graph structure reconstruction loss $\mathcal{L}^{t}_{st}(G)$ at the $t$-th diffusion step as:
\begin{align}
\scriptstyle
\!\! \!   \mathcal{L}^{t}_{ty}(G) &\scriptstyle= \mathrm{CrossEntropy}(\mathbf{h}^t_{ty} \mathbf{W}_e^T, E),\label{eq:ty} \\\label{eq:st}
\scriptstyle\!\!\!\mathcal{L}^{t}_{st}(G) &\scriptstyle= \frac{2}{(m-1)^2}\sum  \limits_{i=1}^{m-1}\sum \limits_{j=i+1}^{m}(\mathrm{MLP}({\mathbf{h}^t_{st}}_i\|{\mathbf{h}^t_{st}}_j)-a_{ij})^2.
\end{align}
The objective of $\mathcal{L}^{t}_{ty}(G)$ in Eq.~(\ref{eq:ty}) is to reduce the difference between the ground truth $E$ and $\mathbf{h}^t_{ty} \mathbf{W}_e^T \in \mathbb{R}^{m\times M}$, which represents the probabilities of each node belonging to each event type.
It is worth noting that $\mathcal{L}^{t}_{ty}(G)$ offers a simplified version of the training objective outlined in Eq.~(\ref{eq:l_e2e}), and empirically improves the quality of the generated schemas.  Meanwhile, the objective of $\mathcal{L}^{t}_{st}(G)$ in Eq.~(\ref{eq:st}) aims to predict the probability of a directed edge from node $i$ to node j and fit their adjacency matrix value $a_{ij}\in \mathbf{A}$.  Finally, we obtain the model by minimizing the following loss:\vsa  
\begin{equation}\label{eq:Dloss}
\footnotesize
\mathcal{L} = \sum \nolimits_{G\in\mathcal{G}} \sum \limits_{t=1}^{T}\mathcal{L}^{t}_{ty}(G)+\lambda\mathcal{L}^{t}_{st}(G),
\end{equation}
where $T$ denotes the total diffusion steps and $\lambda$ is a constant to balance the two objectives. 
When training our model, we randomly select a few instance graphs and then sample a diffusion step $t$ for each of these graphs. We then minimize Eq.~(\ref{eq:Dloss}) to update the model's weights until it converges.


\subsection{Schema Generation}
\label{sec:inference}

We start the schema generation procedure from $\tilde{\mathbf{h}}^T_{la}\in \mathbb{R}^{m\times d}$, which are sampled from Gaussian noise.  We then compute its shared representation $\tilde{\mathbf{h}}^{t}_{sh}$ and the node type representation $\tilde{\mathbf{h}}^{t}_{ty}$ at the $t$-th diffusion step reversely:
\begin{align}
    \tilde{\mathbf{h}}^{t}_{sh} & = \mathrm{E}_{sh}(\tilde{\mathbf{h}}^t_{la}+\mathbf{W}_{pos}+\textsc{EMB}_{s}(t)), \\
\label{aba_2}
\tilde{\mathbf{h}}^{t}_{ty} &= \mathrm{E}_{ty}(\tilde{\mathbf{h}}^{t}_{sh}),
\tilde{\mathbf{h}}^{t-1}_{la} = \tilde{\mathbf{h}}^{t}_{ty}, t=T,\ldots, 1.
\end{align}
After $T$ denoising steps, we obtain the final representation $\tilde{\mathbf{h}}^0_{sh}$, $\tilde{\mathbf{h}}^{0}_{ty}$, and compute $\tilde{\mathbf{h}}^{0}_{st}=\mathrm{E}_{st}(\tilde{\mathbf{h}}^{0}_{sh})$.  

Next, we apply the node type representation $\tilde{\mathbf{h}}^{0}_{ty}$ and the structure representation $\tilde{\mathbf{h}}^{0}_{st}$ to generate the schema.  First, with $\tilde{\mathbf{h}}^0_{ty}=[\tilde{\mathbf{h}}_{ty}^1, \ldots, \tilde{\mathbf{h}}_{ty}^m]\in \mathbb{R}^{m\times d}$, 
we obtain each event's type $e_i\in \tilde{E}$ by assigning the event type whose embedding is nearest to $\tilde{\mathbf{h}}_{ty}^i$ as: 
\begin{equation}
    e_i = \mathop{\arg\min}\limits_{e_j \in \Phi}(\|\tilde{\mathbf{h}}_{ty}^{i}-\textsc{EMB}_{e}(e_j)\|).
\end{equation}
Second, with $\tilde{\mathbf{h}}^0_{st}=[\tilde{\mathbf{h}}_{st}^1, \ldots, \tilde{\mathbf{h}}_{st}^m]\in\mathbb{R}^{m\times d}$, we predict the directed edge from node $i$ to node $j$ where $i<j$ by using a pre-trained classifier $\mathrm{MLP}$ trained via Eq.~(\ref{eq:st}) as follows:
\begin{equation}
    \beta_{ij}=\left\{
    \begin{aligned}
    1, & \ \ \mathrm{MLP}(\tilde{\mathbf{h}}^i_{st}\|\tilde{\mathbf{h}}^j_{st})) > \mathrm{\tau} \\
    0, & \ \ \mathrm{otherwise},
    \end{aligned}
    \right.,
\end{equation}
where $\mathrm{\tau}$ is a threshold to determine the final edges and $\beta_{ij} \in \tilde{\mathbf{A}}$ is the adjacency matrix value of the generated schema. We generate the schema from the reconstructed event sequence $\tilde{E}$ and adjacency matrix $\tilde{\mathbf{A}}$, and remove $\textsc{PAD}$ type events and the edges associated with them and derive the final schema $S$.

\section{Experiments}

\subsection{Datasets}
\label{datasets}
We conduct experiments to evaluate our model in three IED bombings datasets~\cite{li-etal-2021-future,jinetal2022event}. Each dataset associates with a distinct complex event type: \textit{General IED}, \textit{Car bombing IED}, and \textit{Suicide IED}.
Taking the complex event type \textit{Car bombing IED} as an example, to construct the corresponding dataset, we need to build an instance graph set, where each instance graph describes a complex event, e.g., \textit{Kabul ambulance bombing}.
\citet{li-etal-2021-future} first identify some complex events related to the complex event type based on Wikipedia. 
Then, each instance graph is constructed from the reference news articles in Wikipedia pages related to the complex event. Specifically, \citet{li-etal-2021-future} utilized the state-of-the-art information extraction system RESIN~\cite{wen-etal-2021-resin} to extract atomic events, represented as event types, and their temporal relations from news articles, and finally obtained the instance graph set. 
Next, a human curation is performed to ensure the soundness of the instance graphs~\cite{jinetal2022event}. We utilize the released curated datasets for our experiments and follow previous work~\cite{jinetal2022event} to split the data into train, validation, and test sets. The statistics of the three datasets are summarized in Table \ref{table:data_stats}.

\begin{table}[htb]
    \resizebox{\linewidth}{!}{
    \centering
    \begin{tabular}{c|c|c|c}
    \toprule
    \textbf{Datasets} & General-IED & Car-IED & Suicide-IED \\
    \midrule
    \textbf{train/val/test instance graphs}
    & 88/11/12 
    & 75/9/10
    & 176/22/22 \\
    \textbf{Avg e nodes/ee links per graph}
    & 90.8/212.6
    & 146.5/345.7
    & 117.4/245.2 \\
    \bottomrule
    \end{tabular}
    }
    \caption{
    The statistics for the three datasets. ``e'' and ``ee'' denote event and  event-event, respectively.
    }
    \label{table:data_stats}
\end{table}

\subsection{Baselines}
We compare our method with the following strong baselines:
\begin{compactitem}
    \item Temporal Event Graph Model~(\textbf{TEGM})~\citep{li-etal-2021-future}: TEGM
is based on an autoregressive method that step-by-step generates event and edges between newly generated event and existing events and subsequently uses greedy decoding to obtain the schema, starting from a specially predefined \textsc{START} event.
    \item Frequency-Based Sampling~(\textbf{FBS})~\citep{jinetal2022event}: FBS first counts the occurrence frequency of edges between two event types in the train set. Then the schema is constructed in which each node corresponds to one event type, and initially, the schema does not have any edges. After that, FBS samples one pair of event types based on the occurrence frequency of edges and adds an edge between the corresponding nodes into the schema. The process is repeated until the newly added edge resulting in a cycle in the schema.
    \item \textbf{DoubleGAE}~\cite{jinetal2022event}: DoubleGAE generates an event graph based on DVAE~\cite{zhang2019d}. They first use a directed GCN encoder to obtain the mean and variance of the event graph's latent variables, and then according to the sampled latent variables to recover the event graph in an autoregressive paradigm, similar to TEGM. Finally, they obtain the schema by feeding the hidden variables sampled from Gaussian noise into the model.
\end{compactitem}

\subsection{Experimental Setup}
{\bf Quantitative metrics}. We train our model in the train set for a given dataset and then generate the schema according to Sec.~\ref{sec:inference}.
To evaluate the quality of the schema, we compare the schema with the instance graphs in the test set using the following metrics:\\
(1)~\textit{Event type match}. We compute the set of event types in the generated schema and the set for a test instance graph and compute the F1 score between the two sets to see whether our schema contains the event types in the real-word complex events.\\
(2)~\textit{Event sequence match}. We compute the set of event sequences with a length 2 (or 3) in the generated schema, as well as the set for a test instance graph, and compute the F1 scores between the two sets to measure how the schema captures \eat{repeating} substructures in the test instance graphs.


\eat{
Note that we follow the rule in DoubelGAE~\cite{jinetal2022event} to make sure that the comparison is fair, where we generate the same number of candidate graphs and report the one with the best score as the final result.
}

Note that we calculate the average values of each metric above between the generated schema and each instance graph in the test set as the final results. We generate a set of candidate schemas\eat{$\mathcal{S}$} and test their performance on the validation set, and select the best-performing one as the final schema\eat{$S$} for the focused complex event type. 

{\bf Implementation Details.} For our DEGM, the representation dimension $d$ is 256.  The number of encoder layers, $l$, is set to 4.  The graph structure reconstruction loss weight $\lambda$ is 1, and the edge classification threshold $\tau$ is 0.8. The learning rate is 1e-4 and the number of training epochs is 100.  All hyperparameters are chosen based on the validation set. We select the best checkpoint, and the best-performing schema on the validation set according to the event type match~(F1) metric. The maximum number of graph nodes $m$ is 50, and the number of our candidate schema is 500 following~\citet{jinetal2022event}. The event type in DARPA KAIROS ontology is 67. We define the noise schedule as $\overline{\alpha}_t=1-\sqrt{t+1/T}$ following \citet{li2022diffusion} and the total diffusion step $T$ is 100. All the experiments are conducted on Tesla A100 GPU with 40G memory.

\begin{table}[htb]
    \resizebox{\linewidth}{!}{
    \centering
    \begin{tabular}{c|c|c|cc}
    \toprule
    \multirow{2}{*}{\textbf{Datasets}} & \multirow{2}{*}{\textbf{Methods}} & \textbf{Event type} & \multicolumn{2}{c}{\textbf{Event seq match~(F1)}} \\
    & & \textbf{match~(F1)} & \textbf{ $l=2$ } & \textbf{ $l=3$ }\\
    \midrule
    \multirow{4}{*}{General-IED} 
    & TEGM & 0.638 & 0.181 & 0.065  \\
    & FBS & 0.617 & 0.149 & 0.064  \\
    & DoubleGAE & 0.697 & 0.273 & 0.128 \\ 
    & Ours avg
    & {0.726}$_{\pm{0.018}}$ 
    & {0.361}$_{\pm{0.020}}$ 
    & {0.137}$_{\pm{0.009}}$ \\
    & Ours
    & \bf{0.754}$_{\pm{0.008}}$ 
    & \bf{0.413}$_{\pm{0.010}}$ 
    & \bf{0.153}$_{\pm{0.016}}$ \\
    \midrule
    \multirow{4}{*}{Car-IED} 
    & TEGM & 0.588 & 0.162 & 0.044  \\
    & FBS & 0.542 & 0.126 & 0.038  \\
    & DoubleGAE & 0.674 & 0.259 & 0.081 \\ 
    & Ours avg
    & {0.754}$_{\pm{0.008}}$ 
    & {0.413}$_{\pm{0.010}}$ 
    & {0.153}$_{\pm{0.016}}$ \\
    & Ours
    & \bf{0.795}$_{\pm{0.002}}$ 
    & \bf{0.483}$_{\pm{0.030}}$ 
    & \bf{0.357}$_{\pm{0.063}}$ \\  
    \midrule
    \multirow{4}{*}{Suicide-IED} 
    & TEGM & 0.609 & 0.174 & 0.048  \\
    & FBS & 0.642 & 0.164 & 0.036  \\
    & DoubleGAE & 0.709 & 0.290 & 0.095 \\
    & Ours avg
    & {0.744}$_{\pm{0.009}}$ 
    & {0.464}$_{\pm{0.015}}$ 
    & {0.195}$_{\pm{0.052}}$ \\
    & Ours
    & \bf{0.775}$_{\pm{0.005}}$ 
    & \bf{0.534}$_{\pm{0.011}}$ 
    & \bf{0.330}$_{\pm{0.033}}$ \\ 
    \bottomrule
    \end{tabular}
    }
    \caption{
Results of all methods for the three datasets. Our results include the mean and variance under five different random seeds, while other methods' results are from previous work. The best results are in bold.}
    \label{table:main_result}
\end{table}

\eat{
\begin{figure}[!t] 
    \centering
    \includegraphics[width=0.9\linewidth]
    {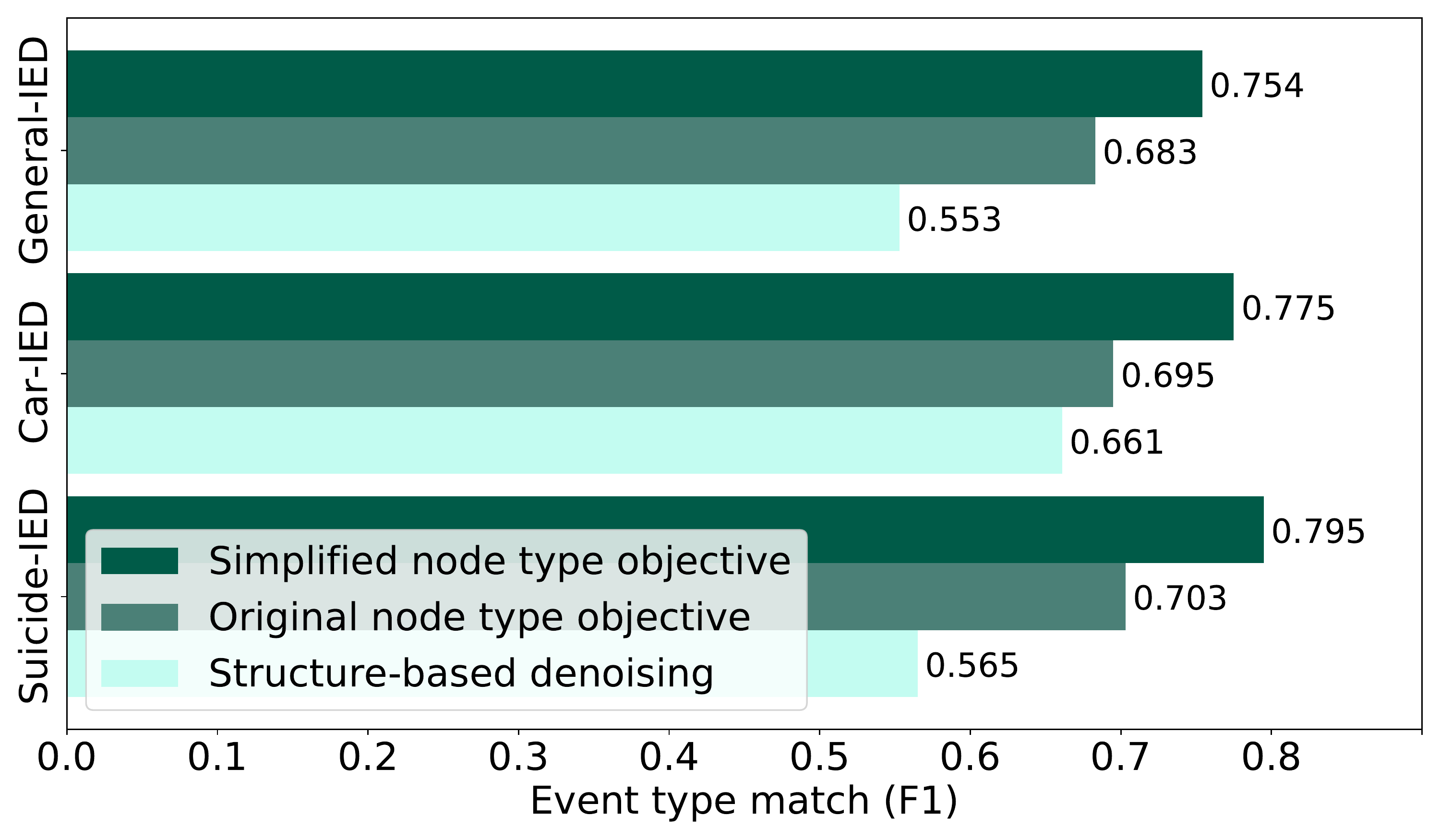}
    \caption{
    We measure the impact of our simplified node type objective and a design choice which means we denoise the schema based on the structure representation. We find that both are crucial for improving the event type match (F1) metric.
    }
    \label{fig:bar}
\end{figure} \vsa 
}

\begin{figure*}[!t] 
    \centering
    \includegraphics[width=1.0\linewidth]{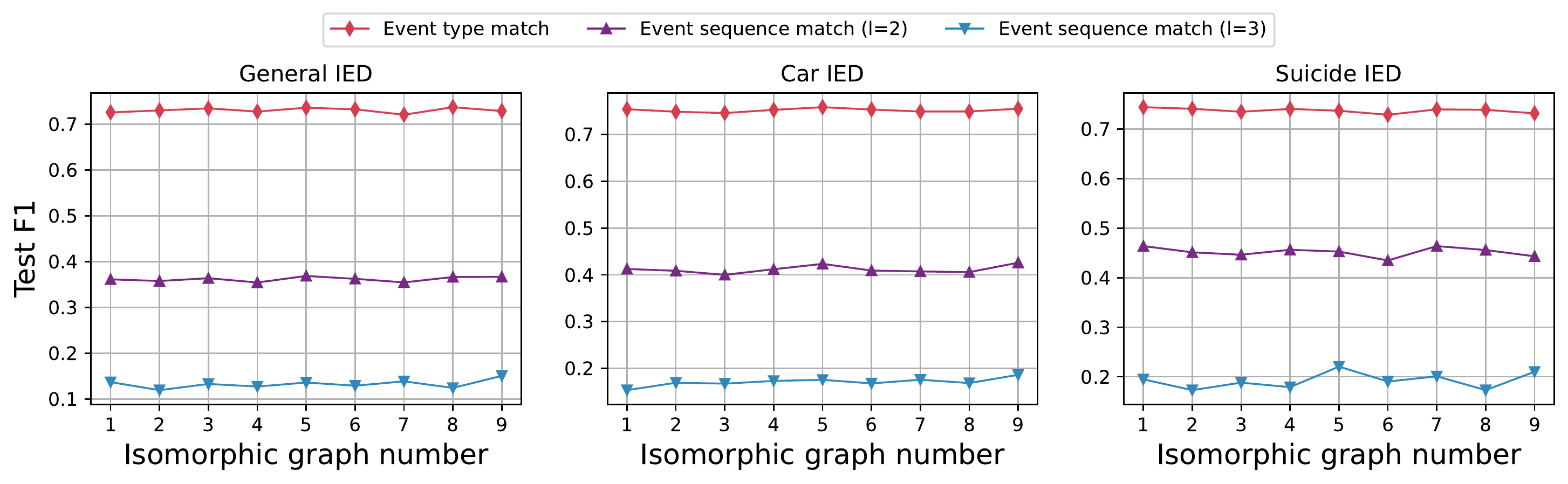}
    \caption{To investigate the impact of topological sorting, we extend the train set by obtaining multiple~(isomorphic graph number) isomorphic instance graphs sorted from one original train instance graph. We train and test our model on the extended dataset. All results are mean values under five different random seeds.}
    \label{fig:random}
\end{figure*}

\begin{figure}[!t] 
    \centering
    \includegraphics[width=1.0\linewidth]
    {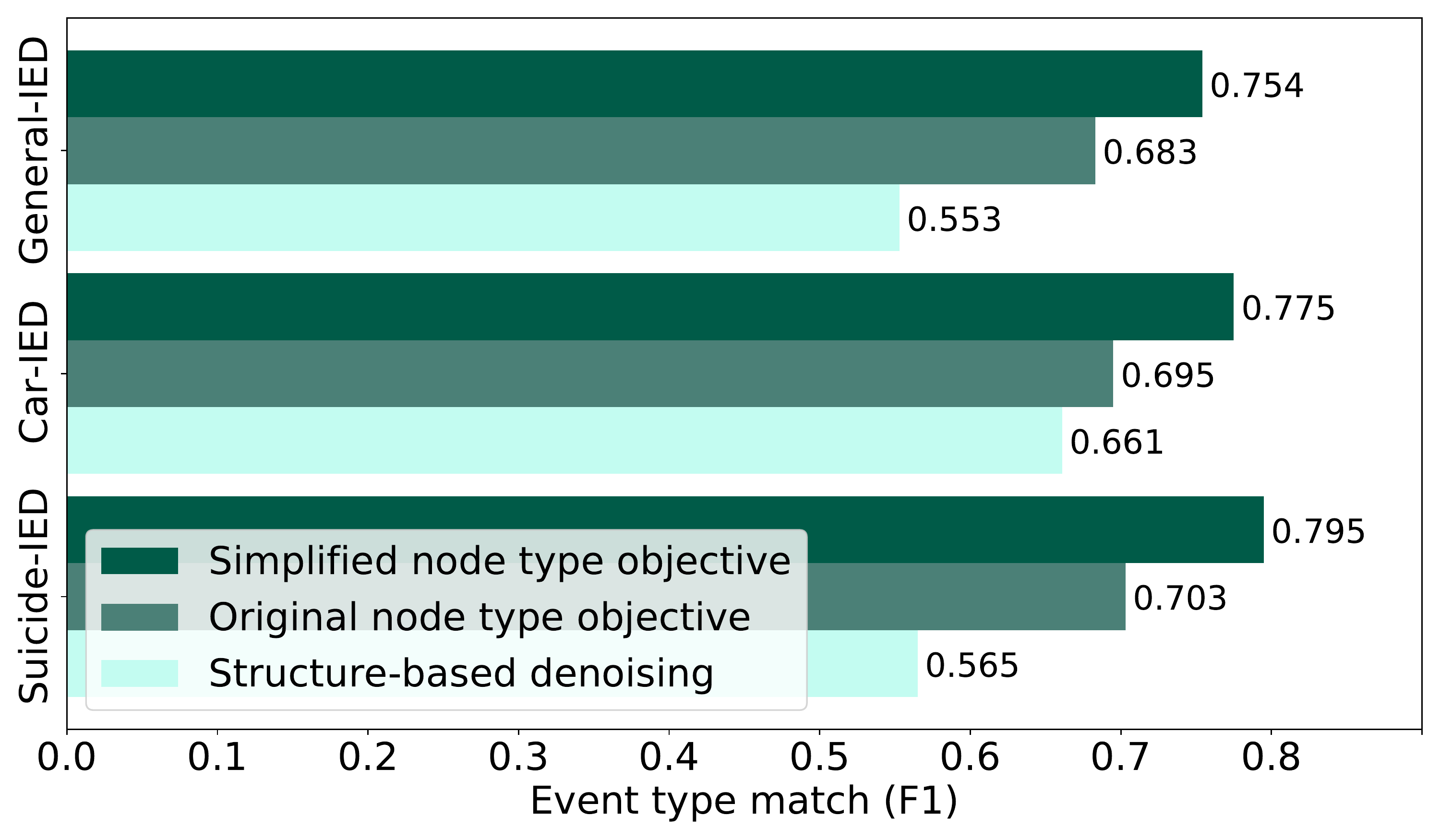}
    \caption{
    We measure the impact of our simplified node type objective and a design choice which means we denoise the schema based on the structure representation. We find that both are crucial for improving the event type match (F1) metric.
    }
    \label{fig:bar}
\end{figure} \vsa 

\subsection{Results and Analysis}

Table~\ref{table:main_result} reports the main results of our model and shows some notable observations:~(1) Our model has achieved significant progress compared to the baselines across three datasets and three metrics;~(2) The average performance of the generated candidate schemas also performs better than previous methods. The reasons for the first observation can be attributed to the ability of our model to iteratively refine the generated schema, enabling the node types and edges between nodes to better match the evolution pattern of the unseen complex events, resulting in superior performance on the test set.
In contrast, Temporal Event Graph Model~(TEGM) can only generate the next event based on the partially generated event graph during training and generation. DoubleGAE has improved this problem by utilizing the encoder structure to capture the global structure of instance graphs. However, DoubleGAE still employs a similar generation procedure as TEGM during schema generation, resulting in a substantial performance gap with our method. Meanwhile, the performance of FBS is much lower than our method, indicating that the heuristic approach is challenging to generate such a schema, demonstrating the necessity for probabilistic modeling for the event graphs. 

For the second observation, we claim that our model is proficient in modeling the distribution of instance graphs. Also, selecting the best-performing schema based on the validation set helps immensely, especially for the event type match~(F1) (l=3) metric. This may be because this metric is more sensitive to the gap between the truth distribution of instance graphs and the modeled distribution, and selecting schema based on the validation set reduces the gap.

\subsection{Ablation Studies}
We verify the importance of our simplified training objective and a design choice while generating the schema through two ablation studies. As shown in Figure \ref{fig:bar}, we can observe that our simplified training objective $\mathcal{L}^{t}_{ty}(G)$ in Eq.~\ref{eq:ty} performs significantly better than the original one Eq.~\ref{eq:l_e2e}.
This may be due to the fact that the original training objective includes three optimization objectives, while ours includes only one. And too many optimization objectives may lead to a larger loss variance, resulting in difficulty in convergence and thus degrading the performance.
At the same time, both training objectives share the same goal: to maximize the model's ability to reconstruct the original event sequence at each diffusion step.

Besides, we also investigate an alternative which we assign $\mathbf{h}^{t-1}_{la}$ as $\mathbf{h}^{t}_{st}$ in Eq.~(\ref{aba_2}) while generating schema. We aim to explore whether it would be better to denoise based on the structure representation $\mathbf{h}^{t}_{st}$. However, this leads to a collapse of the event type match~(F1) metric as in Figure \ref{fig:bar}.
Probably due to the model is trained based on the embedded event sequence to reconstruct the event sequence and its graph structure. Therefore, the model prefers to denoise based on the node type representation $\mathbf{h}^{t}_{ty}$.

\subsection{Impact of Topological Sorting}
Our approach, as well as previous autoregressive graph generation methods, all require a topological sorting of the instance graph to obtain a sorted version of the graph that is not unique. Therefore, we want to investigate whether the model's performance is affected when we train our model with multiple isomorphic instance graphs randomly sorted from one instance graph. Getting $n$ randomly sorted instance graphs from one instance graph is equivalent to expanding the training set $n$ times. We test our model's performance respectively by setting the $n$ range from 1 to 9. As shown in Figure~\ref{fig:random}, however, we observe that training our model on the expanded training set hardly affects the model's performance across all three datasets and three metrics. Indicating that our model captures the evolution pattern of the instance graph based only on one sorted instance graph.



\subsection{Error Analysis and Case Study}

\begin{figure}[!ht] 
    \centering
    \includegraphics[width=0.8\linewidth]
    {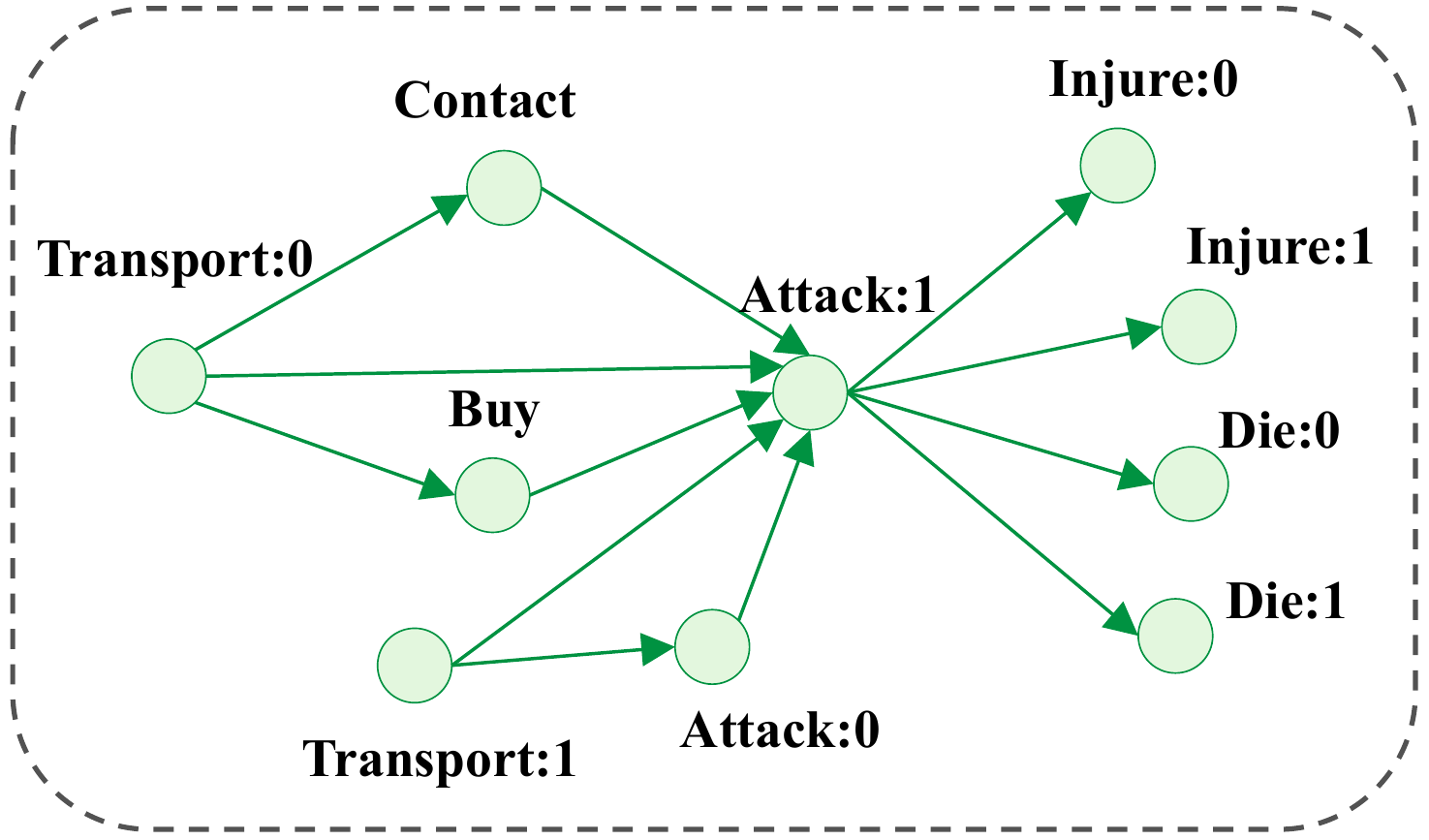}
    \caption{
    A snippet of schema generated by DEGM. 
    }
    \label{fig:case}
\end{figure} \vsa 
In Figure \ref{fig:case}, we present a snippet of the schema generated by our model. From this, we can observe two phenomena: (1) The generated schema contains precise types of atomic events and the common substructures.
(2) The model has a tendency to generate repeated subsequent events and substructures.
The superior performance of our model is revealed by the first phenomenon, which demonstrates its ability to accurately generate both events and substructures. However, the second phenomenon highlights a drawback of the model, which is its tendency to produce duplicate substructures and events. Further analysis revealed that this repetitive structure is caused by a high number of repetitive substructures in the training set, due to the fact that the instance graphs used were extracted from news articles, which can be noisy. As a result, the model learns to replicate these patterns.

\section{Related Work}

According to \citet{jinetal2022event}, event schema induction can be divided into three categories: (1) {\em atomic event schema induction}~\cite{chambers-2013-event,cheung-etal-2013-probabilistic,nguyen-etal-2015-generative,sha-etal-2016-joint,10.1145/3269206.3271674} has focused on inducing an event template, called atomic event schema, for multiple similar atomic events. The template includes an abstracted event type and a set of entity roles shared by all atomic events, while ignoring the relations between events. (2) {\em narrative event schema induction}~\cite{chambers-jurafsky-2008-unsupervised,chambers-jurafsky-2009-unsupervised,jans-etal-2012-skip,rudinger-etal-2015-script,MarkGW-AAAI16,zhu2022generative,gao-etal-2022-improving, gao-etal-2022-mask,DBLP:conf/ijcai/LongCHY22,DBLP:conf/ijcai/YangYDSZZ21}, in contrast, pays attention to the relations between events. In this task, schema is defined as a narrative-ordered sequence of events, with each event including its entity roles. However, complex events in real-world scenarios often consists of multiple events and entities with innerwined relations.

To under such complex events, \citet{li-etal-2020-connecting} incorporate graph structure into schema definition. However, they only consider the relations between two events and their entities. (3) {\em temporal complex event induction task}, recently, \citet{li-etal-2021-future} propose this task in which a schema consists of events, entities, the temporal relations between events, relations between entities, and relations between event and entity~(i.e., argument). Each event and entity is abstracted as an event type or entity type, and each event type contains multiple predefined arguments associated with entities. 
To address this issue, \citet{li-etal-2021-future} generates the schema event by event. Each time an event is generated, the model links it to existing events, expands it with predefined arguments and entities, and links the entities to existing nodes. 
This approach leads to the entities' inability to perceive the events' position, resulting in entities cannot distinguish between events of the same type. Therefore \cite{jinetal2022event} divide the task into two stages: event skeleton generation and entity-entity relation completion. In the first stage, they employ an autoregressive directed graph generation method~\cite{zhang2019d} to generate the schema skeleton, including events and their relations. In the second stage, they expand the schema skeleton with predefined arguments and entities and complete the remaining relations vis a link prediction method VGAE~\cite{kipf2016variational}. 

The above event graph induction methods suffer from error accumulation due to the limitations of the autoregressive schema generation paradigm. To address this issue, we propose DEGM which utilizes a denoising training process to enhance the model's robustness to errors and a schema generationt process to continuously correct the errors in the generated schema.

\section{Conclusions}
We propose Diffusion Event Graph Model, the first workable diffusion model for event skeleton generation.  
A significant breakthrough is to convert the discrete nodes in event instance graphs into a continuous space via embedding and rounding techniques and a custom edge-based loss. The denoising training process improves model robustness. During the schema generation process, we iteratively correct the errors in the schema via latent representation refinement. Experimental results on the three IED bombing datasets demonstrate that our approach achieves better results than other state-of-the-art baselines.

\section*{Limitations}
Our proposed DEGM for event skeleton generation still contains some limitations:
\begin{compactitem}
\item It only considers the problem of event skeleton generation, a subtask of temporal complex event schema induction.  It is promising to explore the whole task, which includes entities and entity-event relations.

\item Perspective from errors found that our model suffers from a tendency to generate correct duplicate substructures.
\end{compactitem}

\section*{Ethics Statement}
We follow the ACL Code of Ethics.  In our work, there are no human subjects and informed consent is not applicable.

\section{Acknowledgments}
The work was fully supported by the IDEA Information and Super Computing Centre (ISCC) and was partially supported by the National Nature Science Foundation of China (No. 62006062, 62176076, 62201576), Natural Science Foundation of GuangDong 2023A1515012922, the Shenzhen Foundational Research Funding (JCYJ20220818102415032, JCYJ20200109113441941), the Major Key Project of PCL2021A06, Guangdong Provincial Key Labo-ratory of Novel Security Intelligence Technologies 2022B1212010005.
\bibliography{anthology,custom}
\bibliographystyle{acl_natbib}

\end{document}